\pdfoutput=1

\documentclass[11pt]{article}

\usepackage[]{EACL2023}

\usepackage{times}
\usepackage{latexsym}
\usepackage[T1]{fontenc}
\usepackage[utf8]{inputenc}
\usepackage{microtype}
\usepackage{inconsolata}

\usepackage{xcolor}
\usepackage{booktabs}
\usepackage{multirow,multicol}
\usepackage{makecell}

\usepackage{graphicx}
\usepackage{caption}
\usepackage{subcaption}
\usepackage{pifont}
\usepackage{enumitem}
\newcommand{\cmark}{\ding{52}}

\newcommand{\add}[1]{\textcolor{black}{#1}}

\title{LED: A Dataset for Life Event Extraction from Dialogs}

\author{
    Yi-Pei Chen\textsuperscript{1}, 
    An-Zi Yen\textsuperscript{2}, 
    Hen-Hsen Huang\textsuperscript{3}, 
    Hideki Nakayama\textsuperscript{1}, 
    Hsin-Hsi Chen\textsuperscript{4}
\medskip
    \\
    \textsuperscript{1} The University of Tokyo, Japan \\
    \textsuperscript{2} National Yang Ming Chiao Tung University, Taiwan \\
    \textsuperscript{3} Academia Sinica, Taiwan
    \textsuperscript{4} National Taiwan University, Taiwan
\medskip
    \\
    \textsuperscript{1}\texttt{ypc@g.ecc.u-tokyo.ac.jp}, \textsuperscript{2}\texttt{azyen@nycu.edu.tw}, \textsuperscript{3}\texttt{hhhuang@iis.sinica.edu.tw}\\ 
    \textsuperscript{1}\texttt{nakayama@ci.i.u-tokyo.ac.jp},
    \textsuperscript{4}\texttt{hhchen@ntu.edu.tw}\\
}

\begin{document}
\maketitle
\begin{abstract}
Lifelogging has gained more attention due to its wide applications, such as personalized recommendations or memory assistance.
The issues of collecting and extracting personal life events have emerged.
People often share their life experiences with others through conversations. 
However, extracting life events from conversations is rarely explored. 
In this paper, we present Life Event Dialog, a dataset containing fine-grained life event annotations on conversational data. 
In addition, we initiate a novel conversational life event extraction task and differentiate the task from the public event extraction or the life event extraction from other sources like microblogs. 
We explore three information extraction (IE) frameworks to address the conversational life event extraction task: OpenIE, relation extraction, and event extraction. 
A comprehensive empirical analysis of the three baselines is established. 
The results suggest that the current event extraction model still struggles with extracting life events from human daily conversations.
Our proposed life event dialog dataset and in-depth analysis of IE frameworks will facilitate future research on life event extraction from conversations.
\end{abstract}

\section{Introduction}
Daily conversation, as a means of communication and switching information, is full of personal information, including personal background, interests and hobbies, connections to other people, and various life events.  
Mining life events lets us better understand a person. 
The extracted life events can be used to construct the personal knowledge base and benefit a variety of downstream tasks, such as lifestyle understanding~\citep{doherty2011passively} and memory assistance~\citep{rahmantowards}.

Previous research on life event extraction mainly focuses on life events from microblogs or social media platforms such as Twitter \cite{li2014major,yen2018detecting,yen2019personal}. 
However, these events from a given fixed passage are static.
In contrast, an event mentioned in a conversation might change its status dynamically throughout the chat. 
Besides, conversations allow participants to interact with each other and gather the information which stimulates participants' interests, revealing people’s general interests in different aspects of information about a life event and expanding additional event information. 
For example, when a person talks about a travel event only with the destination mentioned, the other interlocutor might ask additional information about who they are traveling with, how much the trip cost, and the period and timing of the travel.
Nevertheless, life event extraction from conversations is rarely explored and existing works only detect course or ambiguous event types \cite{eisenberg2020automatic, kao2021convlogminer}. 
The participants and status of events are not recognized, preventing more fine-grained life events analysis and limiting the applications.

We present Life Event Dialog (LED), a dataset with refined life event annotations \add{in English}.\footnote{\url{https://github.com/ntunlplab/LifeEventDialog}} 
We define life events as activities in a person's daily life. 
Following previous works, our life event definition is verb-centered. 
For each event, we annotate three levels of event type from fine-grained to coarse: \textit{Verb}, \textit{Class}, and \textit{Frame}.
Unlike formal writing and social network posts, dialogue is usually in a more flexible and more abstruse style, where the event type is often omitted. 
For example, ``S1: Can I get you some coffee? S2: De-caff.'' indicates an ``order'' event, where the verb ``order'' does not appear in the dialogue. 
Therefore, we also introduce \textit{Explicitness} of an event. 
When the event type cannot be extracted from the dialogue, we manually assign a verb to denote the activity and label the event as an implicit event. 
Besides event types, we annotate \textit{Subject} and \textit{Object} of each event as event participants.
Furthermore, based on the interactive nature of a conversation, more detailed event information is likely to be revealed as the conversation continues. 
People might ask follow-up questions or clarifications in a response that specify the status or attributes of a known event. 
We consider the new supplemental information as the event status change instead of a new event.
To be more specific, we record three aspects of event status: \textit{Polarity}, \textit{Modality}, and \textit{Time}.
These detailed annotations provide more comprehensive information about life events and allow us to track the dynamic event status changes throughout the conversation.

Moving forward from previous research on classifying the types of life events, we introduce the Conversational Life Event Extraction task, which classifies the event type and identifies event participants simultaneously from conversations.
Classifying the event type of a life event is much harder than conventional public event extraction because of the high diversity of life events. 
The form of conversation further adds up to the difficulty of this task. 
For instance, event participants are challenging to identify because they are often in free form, and mentions of the same entity are easily changed throughout the dialogue.
Due to the uniqueness of conversational life event extraction, there has not been a model that specifically tackles this problem. 

In this paper, we examine multiple information extraction (IE) frameworks, including OpenIE, event extraction (EE), and end-to-end relation extraction (RE) models, for this task.
Experimental results show that the existing information extraction models, even the recent models on top of their tasks, still perform poorly in extracting life events from conversations. 
We analyze the strengths and limitations of each model, and urge the development of a better model for Conversational Life Event Extraction.
The contributions of this work are threefold as follows:
\begin{itemize}
    \item We introduce Life Event Dialog (LED) dataset, the first dataset annotated with fine-grained life events in conversations.
    \item We propose a novel task of Conversational Life Event Extraction, stepping forward the event type classification task from previous works.
    \item We explore several IE frameworks on the conversational life event extraction task and offer a thorough analysis of the baselines. 
\end{itemize}

\section{Related Work}
\subsection{Life Event Extraction}
With the rise of social media platforms, people increasingly document their lives online. 
A large amount of personal data is beneficial for applying to lifelogging tasks.
Most life event research collects data from Twitter and contains limited event types. 
\citet{li2014major} gathered tweets with congratulations or condolences replies and proposed a pipeline system to extract 42 major life events like ``getting a job'', ``graduation'', or ``marriage''.
\citet{yen2018detecting} constructed a multi-labeled Chinese tweets dataset with 12 life event types and proposed multiple LSTM models for life events extraction. 
\citet{yen2019personal} built a life event corpus on Chinese tweets focusing on general life events such as dining or visiting a local place, transforming the extracted events into personal knowledge-based facts.
Other than social media posts, the NTCIR14 Lifelog dataset \cite{gurrin2019overview} consists of multimodal lifelogs of images and their metadata. 
They assorted daily activities into 16 categories, but targeted visual lifelog retrieval instead of life event extraction.
Although all concentrate on life events, Conversational Life Event Extraction is distinct from social media or multimodal sources.

\subsection{Conversational Event Extraction}
\citet{li2021reinforcement} designed a task-oriented dialogue system especially for the event extraction task, which differs from our goal of extracting life events from an existing open-domain conversation.
\citet{imani2014evaluating} studied the performance of OpenIE systems on conversations collected from reviews, emails, meetings, blogs, forums, and Twitter. 
Besides the small data size of only a hundred sentences and the dataset not being publically available, their dataset lacks of auxiliary event information such as the event status.

\begin{table*}[t]
\centering
\hspace{-1em}
\tiny
\resizebox{1.0\textwidth}{!}{
\begin{tabular}{cc@{\hskip 0.5mm}p{0.25\linewidth}cllccc}
    \toprule
    D & \multicolumn{2}{c}{Dialogue} & i & Event Types & Participants & P & M & T \\
    \midrule
    \multirow{3}{*}{1} 
        & \textcolor{red}{S1}: & \textcolor{blue}{Bill}, \textcolor{red}{I} must tell \textcolor{blue}{you} the truth. \textcolor{blue}{You} \underline{failed} the English exam again.
    & \multirow{3}{*}{1} 
    & \multirow{3}{*}{\makecell[l]{ \textbf{[Explicit]} \\ \textit{Verb}: failed \\ \textit{Class}: fail \\ \textit{Frame}: Success Act }} 
    & \multirow{3}{*}{\makecell[l]{\textbf{[S]} You \\ \textbf{[O]} English exam}} 
    & \multirow{2}{*}{$+$} 
    & \multirow{2}{*}{$\bigcirc$} 
    & \multirow{2}{*}{before} \\
    \cline{2-3}
        & \textcolor{blue}{S2}: & Ah? Really? That stinks! &  &  &  &  &  & \rule[2.5ex]{0pt}{0pt} \\
    \cline{2-3}\cline{7-9}
        & \textcolor{red}{S1}: & Haha. April Fool's! Did \textcolor{blue}{you} forget what day it is today? &  &  &  
    & $-$ & $\bigcirc$ & before \rule[3ex]{0pt}{0pt}\\
    \midrule
    \midrule
    \multirow{19}{*}{2} 
        & \textcolor{red}{S1}: & Excuse me. \textcolor{red}{I} would like to \underline{purchase} some travelers' checks. 
    & \multirow{3}{*}{1} 
    & \multirow{2}{*}{\makecell[l]{ \textbf{[Explicit]} \\ \textit{Verb}: purchase \\ \textit{Class}: purchase \\ \textit{Frame}: Buy}} 
    & \multirow{2}{*}{\makecell[l]{ \textbf{[S]} I \\ \textbf{[O]} some \\ travelers' checks}} 
    & $+$ & $\triangle$ & now \rule[-3ex]{0pt}{0pt} \\
    \cline{2-3}\cline{7-9}
        & \textcolor{blue}{S2}: & \makecell[cl]{Sure. How much do \textcolor{red}{you} want?} &  &  &  
    & $+$ & $\bigcirc$ & now \rule[4ex]{0pt}{0pt} \\
    \cline{2-9}
        & \textcolor{red}{S1}: & \$5000 and \textcolor{red}{I} want them all in fifties. 
    & 2 
    & \makecell[l]{ \textbf{[Explicit]} \\ \textit{Verb}:  purchase \\ \textit{Class}:  purchase \\ \textit{Frame}: Buy } 
    & \makecell[l]{ \textbf{[S]}  you \\ \textbf{[O]}  \$5000} 
    & $+$ & $\bigcirc$ & now \\
    \cline{2-9}
        & \multirow{2}{*}{\textcolor{blue}{S2}:} & \multirow{2}{*}{OK, here you are. Please \underline{sign} \textcolor{red}{your} name here.} 
    \rule[-4ex]{0pt}{0pt}
    & 3 
    & \makecell[l]{\textbf{[Implicit]} \\ \textit{Verb}: give \\ \textit{Class}: give \\ \textit{Frame}:  Giving} 
    & \makecell[l]{ \textbf{[S]}  S2 \\ \textbf{[O]} S1 \\ \textbf{[O]} \$5000} 
    & $+$ & $\bigcirc$ & now \rule[7ex]{0pt}{0pt} \\
    \cline{4-9}
    & &  
    & \multirow{4}{*}{4} 
    & \multirow{2}{*}{\makecell[l]{\textbf{[Explicit]} \\ \textit{Verb}: sign \\ \textit{Class}: sign \\ \textit{Frame}: Text Creation }} 
    & \multirow{4}{*}{\makecell[l]{ \textbf{[S]}  S1 \\ \textbf{[O]} your name }} 
    \rule[3ex]{0pt}{0pt}\rule[-1ex]{0pt}{0pt}
    & $+$ & $\triangle$ & after \\
    \cline{2-3}\cline{7-9}
    & \textcolor{red}{S1}: & Thank \textcolor{blue}{you}. &  &  &  & $+$ & $\bigcirc$ & now \rule[4ex]{0pt}{0pt}\\
    \\
    \bottomrule
\end{tabular}}
\caption{Two example dialogues with 1 and 4 events, respectively. D: Dialogue ID, i: Event ID.
We display the coreference cluster in \textcolor{red}{red} for S1 and in \textcolor{blue}{blue} for S2. \textit{Verb} of explicit events (extractive) are underlined. 
For each event, we show the event types, participants, and status (\textit{Polarity} (P), \textit{Modality} (M), and \textit{Time} (T)). $+$: positive event, $-$: negative event, $\bigcirc$: actual event, $\triangle$: hypothetical event.}
\label{event_examples}
\end{table*}

\subsection{Life Event Extraction from Conversation}
Works by \citet{eisenberg2020automatic} and \citet{kao2021convlogminer} are the most related works to ours. 
\citet{eisenberg2020automatic} collected conversations from a podcast and classified event features by SVM. Their event annotations only include the event tokens and lack other event information.
\citet{kao2021convlogminer} also constructed a dataset from DailyDialog \cite{li2017dailydialog}, but they only annotated the frame name for each event.
Both works also aimed at extracting personal life events from conversations, yet their proposed datasets only contain plain event annotations. 
In contrast, our LED dataset has more comprehensive annotations, including participants, status, event category, and the coreference clusters of participants.

\section{Life Event Dialog}
In this paper, we define life events as daily life activities, personal habits, life experiences, or personal information of the interlocutors or related people. 
On the other hand, personal feelings or preference, public issues, and general knowledge are not considered life events in our dataset. 

\subsection{Event Schema}
\label{sec:event_schema}
\noindent \textbf{Event Type:} We define three granularities of event type: \textit{Verb}, \textit{Class}, and \textit{Frame}. We also labeled the \textit{Explicitness} based on whether \textit{Verb} can be extracted from the dialogue.
\begin{itemize}
    \item \textit{Explicitness} (E) is determined by whether a verb exists in the dialogue that triggers an event. 
    If no explicit verb exists in the dialogue, but an event is recognized and labeled by annotators, we consider it as an implicit event. 
    See Dialogue 2 Event 3 in Table~\ref{event_examples} for an example.
    \item \textit{Verb} is a verb event trigger, which might be a span extracted from the dialogue (explicit event) or abstractly written by annotators (implicit event). 
    \item \textit{Class} is the fine-grained event type determined by the lemma of \textit{Verb}. 
    \item \textit{Frame} is the coarse event type selected from FrameNet \cite{fillmore2002framenet} by annotators. This event type is also used in previous works \cite{yen2019personal,eisenberg2020automatic,kao2021convlogminer,wang-etal-2020-maven}. 
\end{itemize}
Note that \textit{Frame} and \textit{Class} are not one-to-one mappings. For example, \textit{Class} ``get'' could belong to \textit{Frame} ``Possession'', ``Receiving'', or ``Giving''. In LED, each \textit{Class} belongs to 1.25 \textit{Frame} on average.

\noindent \textbf{Participant:}
We label the span for \textit{Subject} (S) and \textit{Object} (O). 
In a conversation, the same S/O entity might appear recurrently in different mentions, therefore, we also include the coreference cluster ID for S/O as their entity ID. 

\begin{table}[t]
    \centering
    \resizebox{\columnwidth}{!}{
    \begin{tabular}{lrrrrr}\toprule
    & \# Dialogs & U & Evt &Unique Evt \\\midrule
    Train &858 &3,823 &5,529 &1,856 \\
    Valid &75 &349 &593 &179 \\
    Test &70 &313 &426 &151 \\
    \midrule
    Total &1,003 &4,485 &6,548 &2,186 \\
    \bottomrule
    \end{tabular}
    }
    \caption{Dataset Statistics. 
    The number of utterances (U) is the number of training instances (a training instance is an utterance with its dialogue history), and the number of events (Evt) is the cumulative number of events of a training instance. Also, we consider events with same event types and participants as the same event (Unique Evt), which might have different event status.}
    \label{tab:dataset_stat_dialog}
\end{table}

\noindent \textbf{Status:} Three event properties that might change dynamically throughout the dialogue are recorded, including \textit{Polarity}, \textit{Modality}, and \textit{Time}.  
\begin{itemize}
    \item \textit{Polarity} (P) is a binary class of whether an event happens (positive) or does not happen (negative). In some conversations, a life event is specifically expressed in a negative form. Given an utterance, ``You did not invite me to the party.'' We consider the negativity in this sentence as a strong indication of a particular event rather than a random event that doesn't happen. Moreover, an event might change its \textit{Polarity} as the conversation continues. As shown in Dialogue 1 Event 1 in Table~\ref{event_examples}, (You, failed, English exam again) is a positive event in the first two utterances, but after the speaker S1 says it's an April Fool's joke, \textit{Polarity} becomes negative. Therefore, we especially mark the negative event status to keep track of the polarity changes of a life event in the conversation.
    
    \item \textit{Modality} (M) refers to whether an event has happened/is happening (actual), or is mentioned in the dialogue that it will happen in the future (hypothetical), as illustrated in Dialogue 2 Event 1. Note that an event is hypothetical only when indicated in an affirmative sentence and not in a question. For example, (We, have, meeting) in ``We will have a meeting at 9 a.m. tomorrow.'' is a hypothetical event, but (she, call, you) in ``Can she call you back?'' is not.
    
    \item \textit{Time} (T) is labeled as one of ``before'', ``now'', ``after'', ``continuously'', or a specified time span if the time information is explicitly mentioned in the dialogue. \textit{Time} might be related to \textit{Modality}. For instance, one hypothetical event might have \textit{Time} ``after'', waiting for confirmation. After the next utterance reply, the event status would become an actual event with time labeled ``now''. Dialogue 2 Event 4 is an example that changes its status after the last turn is given.
\end{itemize}
The default event status is positive, actual, and happens at now.

\subsection{Annotation Details}
We recruited three annotators with a linguistic degree to annotate the data. 
The dialogue is augmented by one turn at each time, and annotators are asked to label life events for the whole conversation up to the given turns.
To calculate the agreement, we sampled 40 dialogues and asked all annotators to annotate them.
\add{We calculate the agreement on the \textit{Frame} of all positive and actual events in the last turn of each dialogue (the accumulated events in one dialogue). The total number of annotated events are 550.}
The annotation agreement is 0.81, measured by Krippendorff's alpha \cite{krippendorff2011computing}.
For the disagreed cases, we conducted the majority vote or discussed with annotators to re-annotate the event.
\add{The annotation guideline and more annotation details are provided in Appendix~\ref{appendix:guideline}.}

\subsection{Dataset Construction}
\label{sec:dataset_construction}
We sample 1,003 dialogues from the DailyDialog dataset \cite{li2017dailydialog} as the material for life event annotation.
\add{DailyDialog is a multi-turn English dialogue dataset, which contains daily life conversations from various English learning websites. 
The conversations usually focus on a certain topic and under a certain situation, such as a customer finding some goods in a shop.}
We take the five most frequent topics, including Relationship (35\%), Ordinary Life (28\%), Work (20\%), Tourism (9\%), and Attitude \& Emotion (8\%), and annotate four to six utterances of each conversation. 
We include conversations with (73.5\%) and without (26.5\%) events to reflect the real world scenario that not all conversations contain life events.
\begin{table}[]
    \centering
    \begin{tabular}{lrrrrrr}\toprule
    \multicolumn{3}{c}{Unique Event Types} &\multicolumn{3}{c}{Status Change} \\
    \cmidrule(lr){1-3}\cmidrule(lr){4-6}
    \textit{Verb} & \textit{Class} & \textit{Frame} &P &M &T \\
    \midrule
    695 &371 &175 &26 &58 &117 \\
    \bottomrule
    \end{tabular}
    \caption{The number of unique categories in each event type and the number of times when an event changes one of its status.}
    \label{tab:event_stats}
\end{table}

\begin{figure}
    \hspace{-1em}
    \centering
    \includegraphics[trim={20 25 0 30},clip,width=1.02\columnwidth]{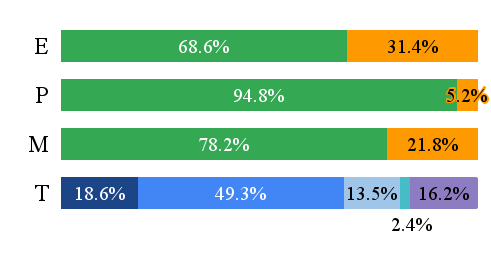}
    \caption{Statistics of \textit{Explicitness} (E) and event status. 
    Green and orange colors stand for explicit/implicit, positive/negative, and actual/hypothetical, for E, P, and M, respectively. Colors of T from left to right are ``before'', ``now'', ``after'', ``continuously'', and the specified time.}
    \label{fig:event_stats}
\end{figure}
Overall, we annotate 2,186 unique life events (Unique Evt) from 4,485 utterances. 
Note that one training instance is an utterance (U) with its dialogue history, and the events of an instance (Evt) would be the cumulative events from the utterance and its dialogue history. 
The statistics of our dataset is shown in Table~\ref{tab:dataset_stat_dialog}. 

For every unique event, the event status might change throughout the conversation. 
We list the number of event status change for P, M, and T, as well as the number of unique event types for \textit{Verb}, \textit{Class}, and \textit{Frame} in Table~\ref{tab:event_stats}. 
The ratio of explicit vs. implicit, positive vs. negative, actual vs. hypothetical events, and the distribution of the T labels are shown in Figure~\ref{fig:event_stats}.

\begin{table*}[!tb]
\centering
\resizebox{\textwidth}{!}{
    \begin{tabular}{cc|cc|cc|cc|c|c}
    \toprule
    LED (Frame) & \% & LED (Class) & \% & ACE2005 & \% & MAVEN & \% & CONLL04 & LiveKB \\
    \midrule
    Statement  & 6.1 & have & 3.9 & Attack & 28.8 & Action & 46.9 & kill & Perception \\
    Perception & 5.3 & go   & 3.8 & Transport & 13.5 & Change & 27.5 & work for & Presence\\
    Motion     & 3.8 & tell & 3.5 & Die & 11.2 & Scenario & 13.4 & organization based on & Using \\
    Request    & 3.2 & hear & 2.8 & Meet & 5.2 & Sentiment & 6.4 & live in & Motion\\
    Ingestion  & 3.2 & see  & 2.8 & End-Position & 4.0 & Possession & 5.7 & located in & Ingestion \\
    \bottomrule
    \end{tabular}
}
\caption{Top 5 event types of our LED dataset compared to other datasets.}
\label{tab:top5_event_types}
\end{table*}
\section{Dataset Analysis}
\label{sec:dataset_analysis}
\subsection{Life Events Distribution}
We list the top five most frequent \textit{Class} and \textit{Frame} among 371 classes and 175 frames in Table~\ref{tab:top5_event_types}, 
from which we can see that either \textit{Class} or \textit{Frame} is sparsely distributed. 
Even the most frequent \textit{Class} accounts for only 3.9\% of all, and the dominant \textit{Frame} makes up only 6.1\%.
The majority event status change is the change of \textit{Time}, which usually happens when people specify the event time. 
The top five implicit event classes are: ``receive'', ``hear'', ``give'', ``invite'', and ``pay''. In contrast, the top explicit event classes are: ``have'', ``tell'', ``go'', ``see'', and ``be''. Three classes (``go'', ``hear'', and ``bring'') are overlapped in top 10 explicit and implicit events classes.

\subsection{Comparison with Event Extraction and Relation Extraction Benchmarks}
\label{sec:datasets_comparison}
\begin{table*}[t]
    \centering
    \resizebox{\textwidth}{!}{
    \begin{tabular}{lllrrrrc}
    \toprule
        Dataset & Task & Source & \# Docs & \# Events & \# Types (Subtypes) & \# Arg Roles & Coref \\
        \midrule
        ACE2005~\citeyearpar{walker2006ace} & EE  & News & 599 & 5,349 & 8 (33) & 35 & \\
        CONLL04~\citeyearpar{roth-yih-2004-linear} & RE  & News & 1,437 & 2,041 & 5 & 4 & \\ 
        LiveKB~\citeyearpar{yen2019personal} & Life EE & Twitter & 25,344 & 15,525 & 137 & 6 \\
        PEDC~\citeyearpar{eisenberg2020automatic} & Life EE & Podcast & 1,038 & 3,664 & 278 & 0 \\
        DiaLog~\citeyearpar{kao2021convlogminer} & Life EE & DailyDialog & 600 & 780 & 21 & 0 \\
        \midrule
        Life Event Dialog  & Life EE & DailyDialog & 1,003 & 2,186 & 175 (371) & 5 & \cmark \\
    \bottomrule
    \end{tabular}}
    \caption{Datasets comparison. EE: Event Extraction, RE: Relation Extaction.}
    \label{tab:datasets_comparison}
\end{table*}
\begin{table*}[!tb]
    \centering
    \begin{tabular}{ccc}
        \toprule
        Framework & Original Output & LED Output \\
        \midrule
        OpenIE & (head, relation, tail) & (S, \textit{Verb} (explicit), O) \\
        RE     & (head, relation, tail) & (S, \textit{Verb}/\textit{Class}/\textit{Frame}, O) \\
        EE     & \makecell[l]{[T span, T type, \\ $A_1$ span, $A_1$ type, $A_2$ span, $A_2$ type, ...]} 
         & \makecell[l]{[\textit{Verb} (explicit), \textit{Class}/\textit{Frame}, \\ S/O, ``subject''/``object'']} \\
        \bottomrule
    \end{tabular}
    \caption{Outputs from OpenIE, RE, and EE frameworks and their mapping to LED output. For EE framework, original output is the span and type of event trigger (T) and the span and type of arguments (A). The T span maps to the span of \textit{Verb} of explicit events; T type maps to \textit{Class} or \textit{Frame} of that event; A span maps to the span of S or O with corresponding ``subject'' or ``object'' string as their A type.}
    \label{tab:framework_outputs}
\end{table*}
Both event extraction (EE) and relation extraction (RE) aim to predict the event type and participant information. 
For EE, each event has a event type (subtype) and argument roles. 
We regard \textit{Frame} and \textit{Class} in LED as the type and sub-types and map \textit{S, O}, and event status (\textit{Polarity}, \textit{Modality}, and \textit{Time}) as the argument roles. 
The RE output is a (head, relation, tail) triple. 
We consider (S, event type, O) in LED as the mapping of a RE triple. 
The major difference between the life events from our LED dataset and the public events from EE/RE benchmarks is the event domain and the distribution of event types. 
Life events in LED belong to a wide variety of categories that are sparsely distributed.
In contrast, current EE and RE benchmarks are often from news reports and focus on certain limited event types.
We compare two EE benchmarks (ACE2005 \cite{walker2006ace} and MAVEN \cite{wang-etal-2020-maven}) and one RE benchmark (CONLL04 \cite{roth-yih-2004-linear}) in Table~\ref{tab:top5_event_types}, demonstrating the distinguishable event type discrepancy on domain and distribution. 

Further, the arguments in EE benchmarks are often a single entity or the head word of a noun phrase, but we often want to keep the informative descriptions of life events, especially for objects. The average object length in LED is 2.95, which is 2.5 times of argument length in ACE2005. 
In addition, a quarter of life events are implicit events, which means 25\% of the event trigger (\textit{Verb}) cannot be found in the text input, whereas all event triggers and arguments are extractable from the given text in EE benchmarks.

\begin{figure}
    \hspace{-1em}
    \centering
    \includegraphics[trim={0 10 0 10},clip,width=1.02\columnwidth]{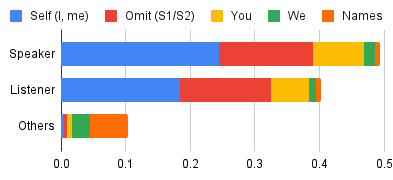}
    \caption{Subject analysis. When S is the speaker, the listener, or others, the mention of S usually belongs to one of the five categories: Self, We, You, Omit, Names.}
    \label{fig:subject_analysis}
\end{figure}
\subsection{Comparison with Life Event Datasets}
LiveKB~\cite{yen2019personal} is a large-scale life event dataset crawled from Chinese Twitter with an event schema similar to ours. 
The major difference between LiveKB and Life Event Dialog derives from the characteristics of a single-person narrative versus interactions between two people. 
In a tweet, the event subject is almost always the author of the tweet if not mentioned.
In contrast, the event subject in a dialogue is half time the speaker, 40\% the listener, and 10\% the others, \add{as shown in Fig.~\ref{fig:subject_analysis}}.
\add{The case of the subject being the listener happens when the event of the listener is told by the speaker, such as ``You are hired by our compan'', ``You get high marks in the exam'', or ``I’m Jame, your neighbor when you lived here last year (indicating the event of the listener living here last year)''.}
\add{Also, besides the case when the speaker themselves being the subject (when the mention is self-referred), the mention of the subject is often omitted (and annotated as S1/S2) or being ``you''. It usually happens when the speaker is confirming an event.} For example, S1: ``Could you please sign this memo?'' S2: ``No problem.'' The event (S2, sign, memo) becomes positive after S2's confirmation. 
These kinds of events that happen after user interactions only appear in our Life Event Dialog data.
There is sometimes an ambiguity regarding the event subject, e.g., S mention ``we'' might refer to only the speaker or both participants in the dialogue.
Further, comparing the top 10 \textit{Frame} in LED and LiveKB, we find that LED has more interactive activities, such as ``Statement'', ``Request'', and ``Acquaintance''.
In contrast, LiveKB activities are more self-centered, like ``Presence'', ``Create'', and ``Buy''. 

Both conversational event extraction datasets, PEDC \cite{eisenberg2020automatic} and DiaLog \cite{kao2021convlogminer}, only annotate event type labels. 
The former is collected from podcast transcripts and focuses on event from life stories told by first-person narrators. The latter classifies events by FrameNet and is also from the DailyDialog.
Our LED has more data, more event types, and additional annotations of argument roles, event status, and coreference clusters, compared with them.

\section{Conversational Life Event Extraction}
We define Conversational Life Event Extraction as the combination of two subtasks: (1) Event Type Classification and (2) Participants Identification.
Given a dialogue $D_u=\{T_1, T_2, ..., T_u\}$ of $u$ turns utterances, we extract $i$ events $E_u^i=\{e_u^1, ..., e_u^i\}$ from $D_u$, where an event $e$ comprises an event type from either \textit{Verb}, \textit{Class}, or \textit{Frame} and spans of participants (S and O). 
We consider an input instance as the concatenation of turns $T_1$ to $T_u$.

\subsection{Frameworks} 
We aim to identify the event type and participants simultaneously. 
By contrast, previous works on life event extraction only dealt with event type prediction. Hence no model specifically tackles our proposed task of conversational life event extraction.
As a result, we examine different information extraction frameworks, including (1) OpenIE, (2) Event Extraction (EE), and (3) Relation Extraction (RE), for this task.
We transform our data schema to fit the original schema of each framework, as shown in Table~\ref{tab:framework_outputs}.
Both OpenIE and RE output (head, relation, tail) triples.
We consider the head and tail to be S and O and relation to be an event type. 
EE outputs the span and type of an event trigger, as well as the span and type of arguments. When converting to our LED schema, the event trigger can be seen as the event type and arguments as participants.
Due to limitations of each framework, the output from each framework is slightly different when adapting to our dataset. 
The major constraint is that OpenIE and EE frameworks can only predict explicit events because both output spans from the input dialogue.

\noindent \textbf{OpenIE:}
OpenIE requires each element in the triplet to be a span from the input, therefore, it is not able to predict event types of \textit{Class} and \textit{Frame}, nor the implicit event which \textit{Verb} is written by annotators.
Also, OpenIE always outputs the whole event triplet, so it can never correctly predict the events without object.
We use Stanford Open IE system~ \cite{angeli2015leveraging} as the OpenIE baseline to extract life event triples.

\noindent \textbf{Relation Extraction:}
RE framework also generates triples as output. 
REBEL \cite{huguet-cabot-navigli-2021-rebel-relation} is selected as the relation extraction baseline, which is based on an autoregressive model BART-large \cite{lewis2019bart}.
Since REBEL is a generation model, it can generate tokens not in the given dialogue and avoid the limitations of OpenIE framework.

\noindent \textbf{Event Extraction:}
Event Extraction framework predicts both spans and their type; 
thus, the implicit events without trigger span can never be predicted.
We choose DyGIE++ \cite{wadden-etal-2019-entity} as our event extraction baseline. 
DyGIE++ is a span-based model with RoBERTa-base \cite{liu2019roberta} backbone, which can perform multi-tasks training on entity recognition, relation extraction, event extraction, and coreference resolution. 

\subsection{Evaluation}
\label{sec:evaluation}
Evaluation metrics vary between frameworks.
We evaluate the output triples from OpenIE and RE using precision (P), recall (R), and micro-F1, following previous works \cite{huguet-cabot-navigli-2021-rebel-relation}. 
We adapt the strict evaluation \cite{taille-etal-2020-lets}, that is, a triple is considered as correct only if the whole triple is exactly the same as the ground truth triplet.
EE results are evaluated by P, R, and F1 of span identification and type classification. 
An event trigger is correctly identified if the span is correct and is correctly classified if the event type is correct. 
An event argument is correctly identified if both the event type and the argument span are correct, and is correctly classified if the argument type is correct. 

We unite evaluation metrics for all frameworks using a lenient evaluation metric. 
For each life event, we first evaluate the event type classification (ET-C) by P, R, and F1.
Then, for those events with correct event type, we evaluate the participants identification by P, R, and F1 of S (S-ID) and O (O-ID F1). 
We also compute BERT Score \cite{bert-score} for the object (O-ID BS), because O in LED are often longer than a single token, unlike in EE/RE datasets (as discussed in Sec~\ref{sec:datasets_comparison}). 

\begin{table*}[t]
\centering
\small
    \begin{tabular}{cccccccccccc}
    \toprule
        Event Type & \multirow{2}{*}{Framework} & \multicolumn{3}{c}{ET-C} & \multicolumn{3}{c}{S-ID} & \multicolumn{4}{c}{O-ID} \\
        Granularity & & P & R & F1 & P & R & F1 & P & R & F1 & BS \\
        \cmidrule(lr){1-1}\cmidrule(lr){2-2}\cmidrule(lr){3-5}\cmidrule(lr){6-8}\cmidrule(lr){9-12}
        \multirow{3}{*}{\textit{Verb}} & OpenIE & 18.5 & 29.1 & 22.6 & 17.3 & 27.2 & 21.1 & 6.5 & 10.2 & 7.9 & 33.5 \\
         & RE & 28.5 & 49.8 & 36.2 & 23.6 & 41.3 & 30.1 & 15.4 & 26.9 & \textbf{19.6} & \textbf{66.2} \\
         & EE & 79.0 & 30.0 & \textbf{43.5} & 64.2 & 24.4 & \textbf{35.4} & 28.4 & 10.8 & 15.6 & 42.0 \\
         & EE + coref & 84.1 & 24.9 & 38.4 & 63.5 & 18.8 & 29.0 & 30.2 & 8.9 & 13.8 & 19.7 \\
        \midrule
        \multirow{2}{*}{\textit{Class}} & RE & 27.6 & 49.3 & 35.4 & 22.9 & 40.8 & 29.3 & 14.7 & 26.3 & \textbf{18.9} & \textbf{64.4} \\
         & EE & 67.8 & 27.7 & \textbf{39.3} & 55.2 & 22.5 & \textbf{32.0} & 26.4 & 10.8 & 15.3 & 42.0 \\
         & EE+coref & 59.2 & 19.7 & 29.6 & 40.8 & 13.6 & 20.4 & 26.8 & 8.9 & 13.4 & 19.7 \\
        \midrule
        \multirow{2}{*}{\textit{Frame}} & RE & 23.4 & 40.4 & 29.6 & 16.3 & 28.2 & 20.7 & 12.0 & 20.7 & \textbf{15.1} & 61.1 \\
         & EE & 58.6 & 23.9 & \textbf{34.0} & 46.0 & 18.8 & \textbf{26.7} & 26.4 & 10.8 &  15.3 & 40.2 \\
         & EE+coref & 57.4 & 12.7 & 20.8 & 57.4 & 12.7 & 20.8 & 21.3	& 4.7 & 7.7 & \textbf{64.0} \\
    \bottomrule
    \end{tabular}
\caption{Result on explicit events across different frameworks evaluated by our lenient evaluation. ET-C: Event Type Classification, S-ID: Subject Identification, O-ID: Object Identification, BS: BERT Score.}
\label{tab:results_united}
\end{table*}

\subsection{Analysis}
Table~\ref{tab:results_united} presents the result of employing each framework on explicit life event extraction, suggesting that the EE framework works the best on event type classification (ET-C) and subject identification (S-ID) over different granularities of event type.
We think the graph-based EE model (DyGIE++) can better capture critical entities and their interactions for event type and S.
The other thing we can benefit from DyGIE++ is that it is compatible with the coreference training, so we can make use of our annotations on participants' coreference clusters. 
However, we are surprised to find that the additional coreference training does not help. 
We suspect that a large amount of examples of the same mention referring to different entities in a dialogue confuse the coreference training. For example, the same subject mention ``I'' might refer to S1 or S2 in different events. 

As for object identification (O-ID), the RE framework gets the top. 
We can see from Table~\ref{tab:results_united} that the bottleneck of Conversational Life Event Extraction is on O-ID, whose F1 score is much lower than the ET-C and S-ID. 
The reason might be the high variance of object mentions. 
We think the best performing RE model (REBEL), an autoregressive model based on a large pretrained language model, is better at copying a sequence of input for O, therefore, can get the best result on O-ID. 
We also found that REBEL often generates repeated output and has higher recall (R) than precision (P), in contrast to DyGIE++, which gets a higher P than R. 

\begin{table}[t]
\centering
\small
\resizebox{\columnwidth}{!}{
\begin{tabular}{llr@{\hskip 0.5em}rr@{\hskip 0.5em}rrrr}\toprule
\multirow{2}{*}{\makecell[c]{Event Type\\ Granularity}} &\multirow{2}{*}{Data} &\multicolumn{2}{c}{ET-C} &\multicolumn{2}{c}{S-ID} &\multicolumn{2}{c}{O-ID} \\
 & &(F1 &$\Delta$) &(F1 &$\Delta$) &(F1 &BS) \\
 \cmidrule(lr){1-1}\cmidrule(lr){2-2}\cmidrule(lr){3-4}\cmidrule(lr){5-6}\cmidrule(lr){7-8}
 \multirow{2}{*}{Verb} &E &36.2 &\multirow{2}{*}{-6.3} &30.1 &\multirow{2}{*}{-9.4} &19.6 &66.2 \\
&E+I &29.9 & &20.7 & &13.9 &57.7 \\
 \midrule
\multirow{2}{*}{Class} &E &35.4 &\multirow{2}{*}{-7.0} &29.3 &\multirow{2}{*}{-8.4} &18.9 &64.4 \\
&E+I &28.4 & &20.9 & &12.0 &57.9 \\
 \midrule
\multirow{2}{*}{Frame} &E &29.6 &\multirow{2}{*}{-5.3} &20.7 &\multirow{2}{*}{-4.3} &15.1 &61.1 \\
&E+I &24.3 & &16.4 & &12.4 &58.3 \\
\bottomrule
\end{tabular}
}
\caption{Event extraction with (E+I) and without (E) implicit events by RE framework.} 
\label{tab:results_implicit}
\end{table}
\begin{table}[t]
\centering
\small
\resizebox{\columnwidth}{!}{
    \begin{tabular}{ccrrrr}
    \toprule
        \makecell[c]{Event Type \\ Granularity} & Framework & \makecell[c]{ET-C \\ (F1)} & \makecell[c]{S-ID \\ (F1)} & \multicolumn{2}{c}{\makecell[c]{O-ID \\ (F1\textcolor{white}{xxxx}BS)}}\\
        \cmidrule(lr){1-1}\cmidrule(lr){2-2}\cmidrule(lr){3-3}\cmidrule(lr){4-4}\cmidrule(lr){5-6}
        \multirow{3}{*}{\textit{Verb}} & OpenIE & 37.6 & 37.6 & 12.5 & 36.8 \\
         & RE & 25.6 & 22.8 & 15.9 & 50.0 \\
         & EE & 21.7 & 21.7 & 0.0 & 8.0 \\
        \midrule
        \multirow{2}{*}{\textit{Class}} & RE & 22.5 & 22.5 & 12.5 & 44.3 \\
         & EE & 21.7 & 21.7 & 0.0 & 8.0 \\
        \midrule
        \multirow{2}{*}{\textit{Frame}} & RE & 0.0 & 0.0 & 0.0	& 51.6 \\
         & EE & 0.0 & 0.0 & 0.0 & 0.0 \\
    \bottomrule
    \end{tabular}
}
\caption{Zero-shot result on explicit events across different frameworks.}
\label{tab:results_zeroShot}
\end{table}

For the three event type granularities, \textit{Verb} is the easiest to predict, and \textit{Frame} is the most challenging. 
The result in Table~\ref{tab:results_united} shows a consistent decreasing trend from \textit{Verb}, \textit{Class}, to \textit{Frame} across all frameworks. 
For ET-C, the gap from \textit{Verb} to \textit{Class} (RE: -0.8, EE: -4.2) is smaller than from \textit{Class} to \textit{Frame} (RE: -5.8, EE: -5.3).
This is intuitive because \textit{Verb} and \textit{Class} are more similar. 
The drastic drop on \textit{Frame} demonstrates the difficulty of inferring the frame name from the dialogue.

RE is the only framework among the three that can deal with implicit events. 
The implicit events account for 31.4\% of all events; hence, we further analyze their impact. 
Table~\ref{tab:results_implicit} shows the results from the RE framework with and without implicit events. 
Despite event type granularities, the results drop after adding implicit events. 
Particularly, when the event type is of \textit{Verb} or \textit{Class}, the negative effect of implicit events is significant (see the $\Delta$ column). 
The results with implicit events are almost the same as the result of explicit events' frame name prediction.  
In other words, predicting an event type that is not in the input dialogue is extremely difficult, and current models cannot achieve promising results. 

We examine the zero-shot result over the three frameworks, when the testing event types are not seen in training time.
The result is shown in Table~\ref{tab:results_zeroShot}.
OpenIE performs the best for ET-C and S-ID on the setting of \textit{Verb} zero-shot. 
Since OpenIE is a rule-based model and does not need any training, it is better than the models required training for unseen event types.
In addition, for the trained models of RE and EE frameworks, they cannot infer any unseen frame name.
Since life events are broad and not fully covered in our dataset, developing models that can extract unseen event types remains an essential research question.

\section{Conclusion}
This work presents Life Event Dialog: a comprehensive life event dataset annotated on DailyDialog conversations. 
The main differences between our dataset and previous datasets on personal life event extraction are:
(1) Life Event Dialog is built on top of conversations instead of microblogs like Twitter. 
The interaction between speakers adds dynamics to events, such as information expansion or status modification, and indicates people's general interests in multiple aspects of other's life events.  
(2) Life Event Dialog contains more data, more types, and more fine-grained event annotations compared to other conversational life event datasets.

We propose the Conversational Life Event Extraction task, extending life event extraction tasks from social media to the conversation domain and from event type detection to predicting both event type and participants simultaneously. 
We then carefully examine three information extraction frameworks: OpenIE, relation extraction (RE), and event extraction (EE), for the pilot study on this task.
The result suggests that current top models on three closely related fields cannot perform well in the Conversational Life Event Extraction task.
Improving object identification and implicit event extraction, detecting unseen life events, and keeping track of event status, constitute our future work. 

\section*{Limitations}
Our LED dataset is annotated on DailyDialog. While annotating on another dataset brings some benefits, it also constrains our dataset. 
For instance, our dataset is limited to the top five frequent topics in DailyDialog, which might not be enough to cover all life events in various scenarios. 
Also, DailyDialog only contains conversations between two interlocutors. For a multi-party conversation, the conversational life events extraction would be much more complicated and interesting. 

The other limitation of LED is the size of the dataset. 
Although with more comprehensive annotations of life events, the number of events in our dataset might not be enough for today's data-hungry models. 
There is always room for larger datasets and more annotations. Compared to the entity types in RE like ``person'', ``organization'', ``location'', to name a few. We do not label such sophisticated argument roles but only ``subject'' and ``object''. We leave this part to our future work.
Besides, we only consider up to 6 turn utterances, yet a dialogue might be much longer in real life. 

Lastly, the definition of life events varies from individual to individual, and our definition of life events might not suit everyone's needs. However, our exploration of the zero-shot experiment shows that it is still possible to find unseen events, and a better model for zero-shot life extraction is needed.

\section*{Ethics Statement}
Our Life Event Dialogs dataset is an extension of an existing public dataset DailyDialog, with all speakers being anonymized in the original release. 
In other words, our dataset does not contain any personally identifiable information that would infringe on someone's privacy. 
In this work, we will only release the life event annotations for research purposes. 
The dialogues in DailyDialog will not be included in LED, but one can access the full DailyDialog dataset from the author's website.\footnote{http://yanran.li/dailydialog}

Our dataset is constructed upon a considerable amount of human annotation. We recruited three annotators and paid them a local hourly wage for the time they spent. The annotation period spanned 1.5 months and resulted in 1,003 annotated conversations (including conversations without events).

\section*{Acknowledgements}
This research was supported by the commissioned research (No. 225) by National Institute of Information and Communications Technology (NICT), Japan, and JSPS/MEXT KAKENHI Grant Numbers JP19H04166 and JP22H05015. 
Chen is supported by JST SPRING, Grant Number JPMJSP2108.
This research was also supported by National Science and Technology Council, Taiwan, under grants MOST 110-2221-E-002-128-MY3 and NSTC 111-2634-F-002-023-.

\bibliography{anthology,custom}
\bibliographystyle{acl_natbib}

\appendix
\section{Annotation Guideline}\label{appendix:guideline}
\subsection{Goal}
We want to extract personal life events related to the speaker according to their dialogue, so that we can construct a personal life knowledge base and benefit other downstream tasks.
\subsection{What \textcolor{red}{are} personal life events?}
\begin{enumerate}
    \item The event happens or might happen in the future to the \underline{interlocutor themselves} or \underline{their relatives and friends}.
    \begin{itemize}
        \item Example: ``I went to Salt Lake City on business with Mr. Wang.''
    \end{itemize}
    \item The event must occur \underline{before the dialog} or \underline{before the dialog ends}.
    \item When expressing \underline{personal thoughts} or \underline{feelings}, the context implies life events.
    \begin{itemize}
        \item Example: ``These cookies taste delicious.'' may imply an event that the speaker has eaten cookies.
    \end{itemize}
    \item The life history or personal information of the interlocutor.
    \begin{itemize}
        \item Example: summer vacation, school start, graduation, ``I skipped fourth grade.'', etc, all belong to life experiences. 
        \item Example: ``I live in Taiwan.'', ``I was born in 1980.'' are personal information.
    \end{itemize}
    \item Interlocutor's personal habits.
    \begin{itemize}
        \item Example: ``I usually look at English language websites \underline{every day} and go to my local English Corner \underline{twice a week}.''
    \end{itemize}
    \item If there is no clear sentence describing an event in the conversation, use the \underline{context} to see if a life event occurred before the conversation completes.
    \begin{itemize}
        \item Example: ``S1: What's for supper? S2: Red cooked carp and rape with fresh mushrooms.'' When the dialogue is completed, it can be deduced that the event ``S2 cooked Red cooked carp and rape with fresh mushrooms for dinner'' occurred.
        \item Example: ``S1: I ran a red light? S2: Yes, you did.'', S1 was originally a question, and the answer of S2 affirmed the occurrence of S1 running a red light.
    \end{itemize}
\end{enumerate}
\subsection{What \textcolor{red}{are not} personal life events?
}
\begin{enumerate}
\item Public issues or general knowledge
\begin{itemize}
    \item Examples: news, knowledge, company business related events.
    \item Examples: ``We run a spotless and cockroach-less hotel.'' Events that represent the company's position are not counted.
\end{itemize}
\item Only expressing personal feelings and preferences (related to emotions)
\begin{itemize}
    \item Examples: ``I feel tired,'' ``I think you are cute,'' ``I like Chinese food,'' ``I'm worried about his condition,'' ``I'm tired of going to school,'' etc.
\end{itemize}
\item Expressing personal abilities
\begin{itemize}
    \item Example: ``I can type 80 words a minute.''
\end{itemize}
\item Things that are not guaranteed to happen don't need to be marked as possible future events
\begin{itemize}
    \item Examples: ``Can you wait a little while?'' ``You should go to school tomorrow.''
\end{itemize}
\item ``Ask questions'' and ``express opinions'' are not considered life events of themselves (unless there is an answer response to judge that an event has occurred)
\begin{itemize}
    \item Example: ``S1: Did you go to school yesterday? S2: No, I didn't.'' Only need to mark the event ``S2 did not go to school yesterday'', and do not need to mark the event ``S1 asked S2 a question''.
\end{itemize}
\item A simple description of the environment, people, things, and things is not considered a life event (unless there is an implied life event)
\begin{itemize}
    \item Example: ``That girl standing there is pretty.''
\end{itemize}
\end{enumerate}
\subsection{Event Explicitness}
Events can be classified into \textit{Explicit} or \textit{Implicit} events, depending on whether there is a clear action in the sentence to indicate the occurrence of the event.

\noindent \textbf{Explicit Event}: There exists an explicit action describing a life event. 
\begin{itemize}
\item As long as the Predicate appears in the dialogue that clearly represents the action of the event, it belongs to the \textit{Explicit} event. If there is a verb but it is not clear, please deduce the explicit verb and mark it as \textit{Implicit}.
\begin{itemize}
    \item Example: ``S1: I ran a red light? S2: Yes." $\rightarrow$ Explicit Event: (Subject= S1, Predicate= ran , Object= red light, Time= BEFORE, Polarity= POS, Modality= ACTUAL)
\end{itemize}
\item Object can be missing, for example: ``We'll wait.'' with a clear action (wait).
\item If the life event has been explicitly described, it is not necessary to extend the label to other possible events.
\begin{itemize}
    \item Example: ``Today I played basketball.'' There is no need to mark the event of ``I went to the basketball court.''
\end{itemize}
\end{itemize} 

\noindent \textbf{Implicit Event}: Contexts and situations are required to infer an life event. (As long as the Predicate needs to be deduced, it is considered \textit{Implicit}).
\begin{itemize}
\item Please infer the action most relevant to your life experience based on the dialogue context.
\item A sentence with an ambiguous verb.
\begin{itemize}
    \item Example: ``I want a fillet steak, medium.'' In the context of ordering food, please deduce that the Predicate is ``order'', and mark the event as \textit{Implicit}. $\rightarrow$ Implicit Event: (Subject= I, Predicate= order , Object= fillet steak, medium, Time= NOW, Polarity= POS, Modality= ACTUAL)
\end{itemize}
\item Events implicit in the dialogue.
\begin{itemize}
    \item Example: ``S1 : Can I get you some coffee? S2 : De-caff.'' $\rightarrow$ Implicit Event: (Subject= S2, Predicate= order , Object= De-caff, Time= NOW, Polarity= POS, Modality= ACTUAL)
\end{itemize}
\item Implicit event in a sentence.
\begin{itemize}
    \item ``S1 : You must be exhausted after your long trip from Canada.'' $\rightarrow$ Implicit Event: (Subject= You, Predicate= travel from , Object= Canada, Time= BEFORE, Polarity= POS, Modality= ACTUAL)
\end{itemize}
\item The situations of the dialogue, such as order meals, make phone calls, send things, job interviews, etc.
\begin{itemize}
    \item Example: ``S1 : This is John speaking. S2 : Hi, this is Mary.'' $\rightarrow$ Implicit Event: (Subject= S2(Mary), Predicate= call, Object= S1(John))
\end{itemize}
\item Note: Except for the Predicate of Implicit Event, please use the vocabulary in the sentence for Subject, Predicate, Object, and Time of Explicit Event, and do not create your own vocabulary.
\end{itemize}
\subsection{Format Description}
The annotation for an event includes the following fields: Subject, Predicate, Object, Time, Polarity, Modality.

\noindent \textbf{Subject}: The subject is the word that performs the action. Most subjects are nouns, pronouns, noun phrases or noun clauses.
Subjects are mainly the two interlocutors, but may also be people or things related to life events.

\noindent \textbf{Predicate}: The action of a life event, expressing what the subject did or what happened. Usually a verb, but may also be a preposition (please refer to the example label below).

\begin{itemize}
    \item Predicate needs to indicate a clear action. 
    \begin{itemize}
        \item Example: "I'd like to take the apartment I looked at yesterday.", take means accept, but we know from the above that the interlocutor wants to rent a house, so please mark the more specific action rent as a Predicate.
        \item Example: ``I need a double and three triples.'', need means need, but it can be inferred in the dialogue that the interlocutor wants to book a room, so please mark the action book as Predicate.
        \item Example: ``I'll be right there.'' This sentence means that I will go to a certain store immediately, please do not directly mark (I, be, there), please deduce a more precise action go to from the predicate
    \end{itemize}
    \item When Predicate is a preposition, please mark it according to the following example:
    \begin{itemize}
        \item "with" means "and", which means an event involving more than two people. 
        \begin{itemize}
            \item Example: ``I went shopping with her.'' or ``I went shopping ... with her.'' \\
            $\rightarrow$ Explicit Event 1: (Subject= I, Predicate= went , Object= shopping, Time= BEFORE, Polarity= POS, Modality=ACTUAL, ) \\ 
            $\rightarrow$  Explicit Event 2: (Subject= I, Predicate= went with / with , Object= her, Time= BEFORE, Polarity= POS, Modality= ACTUAL)
        \end{itemize}
        \item Modifies verbs, such as prepositions denoting the destination and means of movement.
        \begin{itemize}
            \item Example: ``I went to San Francisco by plane.'' \\
            $\rightarrow$ Event 1: (Subject= I, Predicate= went to , Object= San Francisco, Time= BEFORE, Polarity= POS, Modality= ACTUAL) \\
            $\rightarrow$ Event 2: (Subject= I, Predicate= went by , Object= plane, Time= BEFORE, Polarity= POS, Modality= ACTUAL)
            \item Example: ``He is on the school volleyball team." \\
            $\rightarrow$ Event: (Subject= He, Predicate= is on , Object= school volleyball team, Time= CONTINUOUSLY, Polarity= POS, Modality= ACTUAL)
            \item Example: ``S1 : Did you hear it on the radio? S2 : Yes." \\
            $\rightarrow$ Event 1: (Subject= S2, Predicate= hear , Object= it, Time= BEFORE, Polarity= POS, Modality= ACTUAL)\\
            $\rightarrow$ Event 2: (Subject= S2, Predicate= hear on , Object= radio, Time= BEFORE, Polarity= POS, Modality= ACTUAL)
        \end{itemize}
        \item If the preposition refers to the time, please mark the time directly in the field of Time.
        \begin{itemize}
            \item Example: ``We ate dinner at 8 pm''\\
            $\rightarrow$ Event: (Subject= We, Predicate= ate, Object= dinner, Time= 8 pm, Polarity= POS, Modality= ACTUAL)
        \end{itemize}
    \end{itemize}
    \item Nested events.
    \begin{itemize}
        \item Example: ``I'm planning to sing a song in front of everybody.'' \\
        $\rightarrow$ Event 1: (Subject= I , Predicate= 'm planning to , Object= sing a song in front of everybody , Time= NOW, Polarity= POS, Modality= ACTUAL) \\
        $\rightarrow$ Event 2: (Subject= I , Predicate= sing , Object= song , Time= AFTER, Polarity= POS, Modality= HYPOTHETICAL) \\
        $\rightarrow$ Event 3: (Subject= I , Predicate= in front of , Object= everybody , Time= AFTER, Polarity= POS, Modality= HYPOTHETICAL) \\
    \end{itemize}
    \item Sentences that describe situations where no event occurred.
    \begin{itemize}
        \item Example: ``John didn't go to the party tonight.'' Predicate does not need to mark negative words (didn't), please mark positive or negative marks in Polarity.
    \end{itemize}
    \item Sentences describe possible future events.
    \begin{itemize}
        \item Example: ``We will have a meeting at 9 am tomorrow.'' Predicate does not need to mark auxiliary verbs that indicate future occurrences (for example: will, is going to), please mark the form of event occurrence in Modality.
    \end{itemize}
    \item Not a predicate of personal life events: think, know, need, want, hope, trust, like, feel.
\end{itemize}

\noindent \textbf{Object}: The object may be a person, thing, or object, expressing the relationship with the Subject through the Predicate. Most are nouns, pronouns, noun phrases or noun clauses.

Please use words that appear in the dialogue as much as possible, and only mark words that are meaningful to the event.
\begin{itemize}
    \item Example: ``I have a hat.'' Do not need to annotate articles (such as ``a", ``the").
    \item Example: ``I made this delicious dinner.'' Do not need to annotate the adjective.
    \item Example: ``I have a problem with my room.'' Supplemental words such as ``with my room'' need to be annotated.
\end{itemize}

\noindent \textbf{Time}: Express the time information of the life event, such as the time or frequency of the event.

If there is a clear description of the time information in the dialogue, for example: yesterday, last week, directly fill in the time information in the sentence.

If there is no clear description, the default time mark can be filled in as follows:
\begin{itemize}
    \item BEFORE : Indicates that the event occurs before the dialog occurs.
    \item NOW : Indicates that the event occurred during the period from the beginning of the conversation to the end of the conversation
    \item CONTINUOUSLY : Indicates that the event has continued to occur from the past to the present (longer duration).
    \item AFTER : Indicates that the event (possibly) happens after the conversation ends.
\end{itemize}
Please infer which label is suitable for filling in according to the dialogue. 

If there is a vague description in the sentence, please fill in the mark that matches the meaning of the adverb of time.
\begin{itemize}
    \item Example: ``I just finished my homework.'' Please fill in NOW for Time.
\end{itemize}

If people use ``after...'' or ``before...'' to describe the occurrence time in the sentence, you can fill in it directly.

\noindent \textbf{Polarity}: Indicates that the life event is positive or negative. The default is POS for positive and NEG for negative.
\begin{itemize}
    \item Example: ``You did not invite me to the party .''\\
    $\rightarrow$ Event 1: (Subject=You, Predicate=invite, Object=me, Time=BEFORE, Polarity= NEG , Modality=ACTUAL)\\
    $\rightarrow$ Event 2: 
    (Subject=You, Predicate=invite to, Object=party, Time=BEFORE, Polarity= NEG , Modality=ACTUAL)
    \item Example: ``I have no money with me.''\\
    $\rightarrow$ Event: (Subject=I, Predicate=have, Object=money, Time=NOW, Polarity= NEG , Modality=ACTUAL)
\end{itemize}

\noindent \textbf{Modality}: Indicates the form of life events, with the following symbols:
\begin{itemize}
    \item ACTUAL: Indicates that the event has occurred before or at the moment when the sentence is spoken.
    \item HYPOTHETICAL: Indicates that the event may happen in the future, but only if there is a clear sentence in the dialogue to affirm or deny that the future will do. Even if the next moment of speaking may happen but has not happened yet, please mark it as HYPOTHETICAL. After adding the next sentence of dialogue, the situation can be deduced that it has happened, and then changed to ACTUAL.
\end{itemize}

\subsection{Coreference Annotation}
Mark all words in the dialogue that point to pronouns in the Event. Mark all the words representing the same thing into the same mention.
\begin{itemize}
    \item Example: ``S1 : Did you eat the cake on the table? S2 : Yes, I ate that.''\\
    $\rightarrow$ Explicit Event: (Subject= I, Predicate= ate, Object= that, ...) \\
    $\rightarrow$ Coref tag: (Subject: (I, S2), Object: (that, cake on the table))
\end{itemize}
\begin{figure*}[t]
    \centering
    \includegraphics[width=\linewidth]{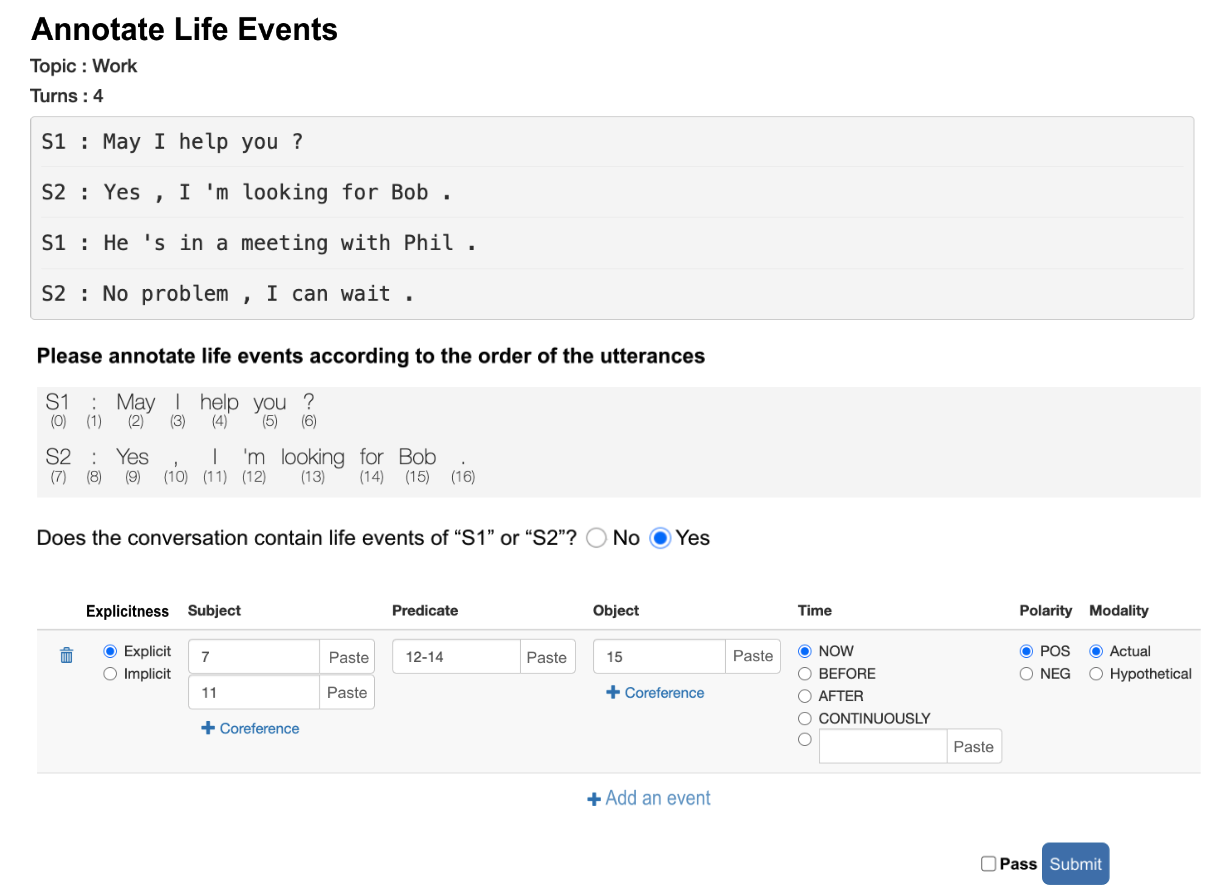}
    \caption{The annotation interface.}
    \label{fig:annotation_interface}
\end{figure*}

\section{Annotation Interface}
Figure~\ref{fig:annotation_interface} shows the annotation interface.
The annotator was first shown the topic of the conversation, the number of turns to annotate, and the full dialogue.
Then, the utterances of the dialogue are displayed turn by turn cumulatively. The example in Fig~\ref{fig:annotation_interface} is the second instance of the dialogue.
The annotators should decide whether the cumulative turns contain life events of the speakers.
If answering ``Yes'', they will add the index of ``Subject'', ``Predicate'' (if it's an explicit event), and ``Object'', and select the event status.

\end{document}